\documentclass{article}
\date{}
\usepackage[english]{babel}

\usepackage[letterpaper,top=1.5cm,bottom=2.5cm,left=2cm,right=2cm,marginparwidth=1.5cm]{geometry}
\providecommand{\keywords}[1]{\textbf{\textit{Keywords:}} #1}
\usepackage{setspace}
\usepackage{amsmath}
\usepackage{graphicx}
\usepackage[table,xcdraw]{xcolor}
\usepackage{caption}
\usepackage{subcaption}
\usepackage[colorlinks=true, allcolors=blue]{hyperref}
\usepackage{rotating}
\usepackage{adjustbox}
\usepackage[normalem]{ulem}
\useunder{\uline}{\ul}{}
\usepackage{lscape}
\usepackage{longtable}

\title{Breast Cancer Diagnosis: A Comprehensive Exploration of Explainable Artificial Intelligence (XAI) Techniques}

\usepackage{authblk}
\author[1]{Samita Bai}
\author[2]{Sidra Nasir}
\author[3]{Rizwan Ahmed Khan\thanks{corresponding author: Rizwan.khan@shu.edu.pk}}
\author[4]{Alexandre Meyer}
\author[5]{Hubert Konik}

\affil[1]{Canadian Institute of Cybersecurity (CIC), University of New Brunswick, Fredericton, New Brunswick, Canada}
\affil[2]{Dipartimento di Informatica, Università di Verona, Italy}
\affil[3]{Department of Computer Science, School of Mathematics and Computer Science, Institute of Business Administration, Karachi, Pakistan}
\affil[4]{LIRIS, CNRS, Université Claude Bernard Lyon 1, 69100 Villeurbanne, France}
\affil[5]{LHC, UMR5516, Université de Saint-Etienne, F-42000 Saint-Etienne, France}

\begin{document}
\maketitle

\begin{abstract}
Breast cancer (BC) stands as one of the most common malignancies affecting women worldwide, necessitating advancements in diagnostic methodologies for better clinical outcomes. This article provides a comprehensive exploration of the application of Explainable Artificial Intelligence (XAI) techniques in the detection and diagnosis of breast cancer. As Artificial Intelligence (AI) technologies continue to permeate the healthcare sector, particularly in oncology, the need for transparent and interpretable models becomes imperative to enhance clinical decision-making and patient care. This review discusses the integration of various XAI approaches, such as SHAP, LIME, Grad-CAM, and others, with machine learning and deep learning models utilized in breast cancer detection and classification. By investigating the modalities of breast cancer datasets, including mammograms, ultrasounds and their processing with AI, the paper highlights how XAI can lead to more accurate diagnoses and personalized treatment plans. It also examines the challenges in implementing these techniques and the importance of developing standardized metrics for evaluating XAI's effectiveness in clinical settings. Through detailed analysis and discussion, this article aims to highlight the potential of XAI in bridging the gap between complex AI models and practical healthcare applications, thereby fostering trust and understanding among medical professionals and improving patient outcomes.
\end{abstract}

\keywords {Breast Cancer Diagnosis, Explainable Artificial Intelligence (XAI), Grad-CAM, LIME, SHAP}

\section{Introduction} \label{introduction}
\begin{figure}[ht!]
    \centering
    \begin{minipage}[b]{0.45\textwidth}
        \centering
        \includegraphics[width=\textwidth]{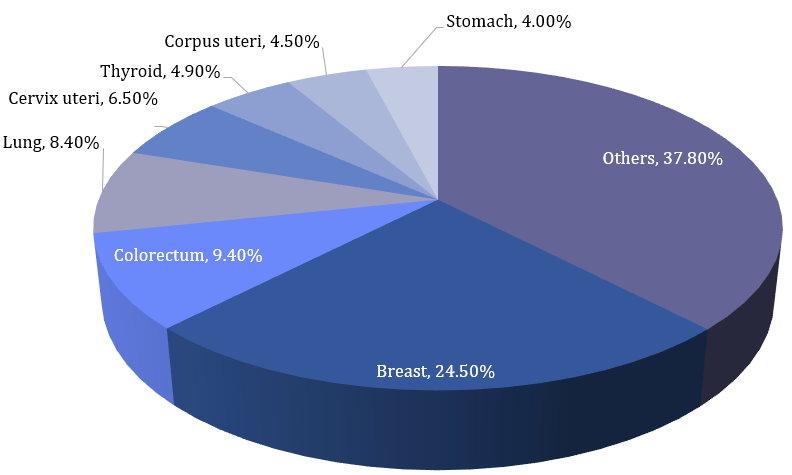}
        \caption{Statistics of occurrence of new cases of different cancers in female patients~\cite{sung2021global}.}
        \label{fig1}
    \end{minipage}
    \hfill
    \begin{minipage}[b]{0.45\textwidth}
        \centering
        \includegraphics[width=\textwidth]{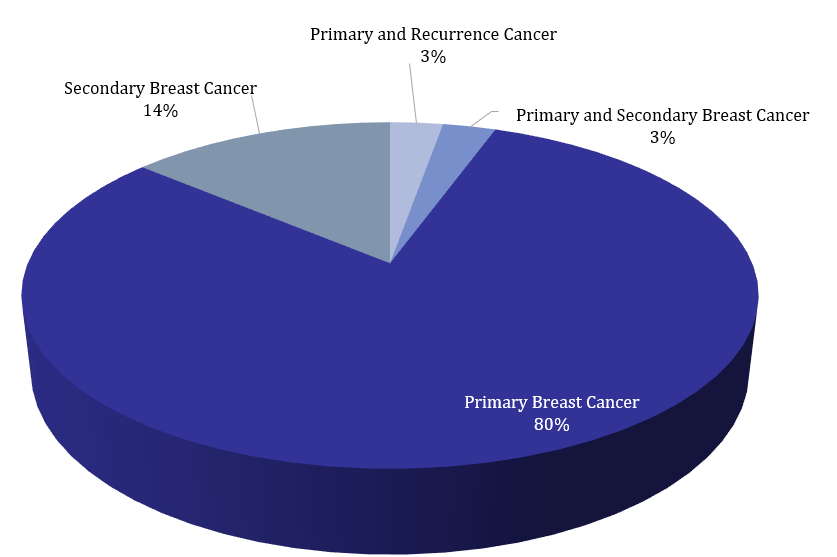}
        \caption{Distribution of breast cancer studies by type.}
        \label{fig13}
    \end{minipage}
\end{figure}

Breast cancer (BC) ranks as the predominant form of cancer in adults worldwide, with an alarming rate of over 2.3 million new cases each year, as reported by the World Health Organization (WHO) 2022~\cite{WHO_breast_cancer}, as shown in Figure~\ref{fig1}. Breast cancer survival rates vary significantly across the globe, with a majority of deaths occurring in low- and middle-income countries. Early detection is crucial as it leads to a clinical cure rate of over 90\%, but this decreases significantly as the cancer progresses~\cite{sung2021global}. Integrating artificial intelligence (AI) techniques into breast cancer diagnosis can improve early diagnosis and patient outcomes. AI systems' ability to analyze large datasets, recognize patterns, and perform predictive analysis can enhance decision-making and personalized care, ultimately improving the precision and effectiveness of diagnosis and treatment strategies~\cite{dileep2022artificial}.

Breast cancer is categorized into three primary groups that reflect the stage of the disease and the complexity of treatment. Primary breast cancer, which makes up 80\% of cases, is found initially in the breast without having spread beyond the breast or the lymph nodes under the arms. Secondary breast cancer, constituting 14\% of cases, involves cancer spreading to other parts of the body, such as the lungs, bones, or liver. The smallest group, which accounts for 6\% of cases, includes those with both primary and secondary characteristics due to recurrence, indicating that the cancer has returned after treatment, either at the original site or elsewhere. The studies reviewed in this article are distributed according to the abovementioned categories, as depicted in Figure~\ref{fig13}.

AI in healthcare optimizes diagnostics and treatments by analyzing vast medical data, including Electronic Health Records (EHRs) and imagery, to identify patterns and predict outcomes accurately. It aids in detecting diseases like cancer in medical images, enhancing diagnoses and patient outcomes. By analyzing medical history and genetic data, AI enables early disease detection and suggests preventive measures, improving patient care. Moreover, by automating routine tasks, AI reduces healthcare costs and streamlines workflows, allowing providers to focus on complex cases and improving patient access to care~\cite{Esteva2019}.

However, some limitations must be considered. One of the limitations is the inadequate diversity of the training data utilized. For instance, if the training data consists of images of breasts from only one ethnic group, it may not accurately identify cancer in breasts from other ethnic groups. Moreover, the systems might lack clarity, posing a challenge for doctors to comprehend the reasoning guiding the system's decision-making process. This lack of transparency could be troublesome in medical contexts, where human expertise and intuition are necessary for accurate diagnosis~\cite{Topol2019}.

To address these limitations, Explainable Artificial Intelligence (XAI) could be a promising approach by offering clear explanations of AI systems' decision-making processes. This clarity increases trust in recommendations for both medical professionals and patients, aiding informed decision-making. Additionally, interpretability helps reduce false positives and negatives, enhancing accuracy and minimizing risks in diagnosis and treatment~\cite{ghassemi2021false, caffo2022explainable, stiglic2020interpretability}. This article provides a comprehensive overview of the state-of-the-art breast cancer detection techniques utilizing XAI techniques such as case-based reasoning (CBR), visualization, and feature importance analysis. 

The paper begins with an introduction, followed by Section~\ref{review_method}, which details the review methodology and protocol employed in this survey. Next, Section~\ref{Datasets}, provides an overview of the datasets and modalities used for breast cancer diagnosis. Subsequently, Section~\ref{XAI} provides an overview of XAI. Afterward, Section~\ref{XAI_techs} discusses various XAI models that are applied to explain Machine Learning (ML) and Deep Learning (DL) models for primary or secondary breast cancer detection and their recurrence, offering insights into the strengths and weaknesses of existing approaches. Furthermore, a comprehensive discussion of the most frequently used XAI techniques is provided in Section~\ref{discussion}. The paper concludes with a discourse on the implications of the findings and future directions in Section~\ref{conc}.

\section{Review Method}\label{review_method}
As previously indicated, this review aims to find and present research on using XAI in breast cancer diagnosis by carefully choosing and evaluating pertinent studies. Briefly, this study attempts to identify the challenges in traditional breast cancer detection and assess the effectiveness of employing XAI techniques to mitigate them. 
The study will examine different XAI methods, including model-agnostic and model-specific approaches, to assess their ability to enhance breast cancer detection accuracy. While model-agnostic XAI aims for interpretability across diverse machine learning models without relying on specific internal structures, model-specific methods tailor explanations to individual model characteristics. Additionally, the study will also investigate how XAI techniques can enhance trust in automated detection systems among healthcare professionals and patients. Lastly, it will examine how XAI affects medical professionals in making decisions about breast cancer diagnosis and possible effects on patient outcomes.

 \subsection{Review Protocol}\label{review protocol}
This study was initiated by developing a thorough review protocol aligned with established methodologies for comprehensive literature surveys. The protocol serves as a guide for the survey process by outlining key components such as the background of the survey, the strategy for literature search, methods for data extraction, and criteria for assessing the quality of included studies. 

\begin{figure}[ht!]
\centering
\includegraphics[width=0.9\textwidth]{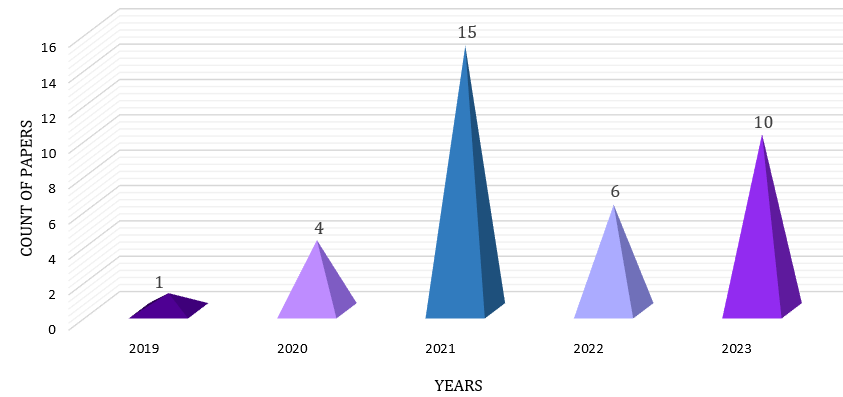}
\caption{The distribution of studies about breast cancer diagnosis using XAI spanning from 2019-2023.}\label{fig12}
\end{figure}

While not strictly adhering to the systematic literature review (SLR) framework \cite{kitchenham2010systematic}, a review protocol ensures clarity, consistency, and transparency in the survey methodology. By delineating predefined procedures for study selection and analysis, the review protocol minimizes potential biases and enhances the reliability of survey findings.

\begin{figure}[htb!]
\centering
\includegraphics[width=16cm, height=7cm]{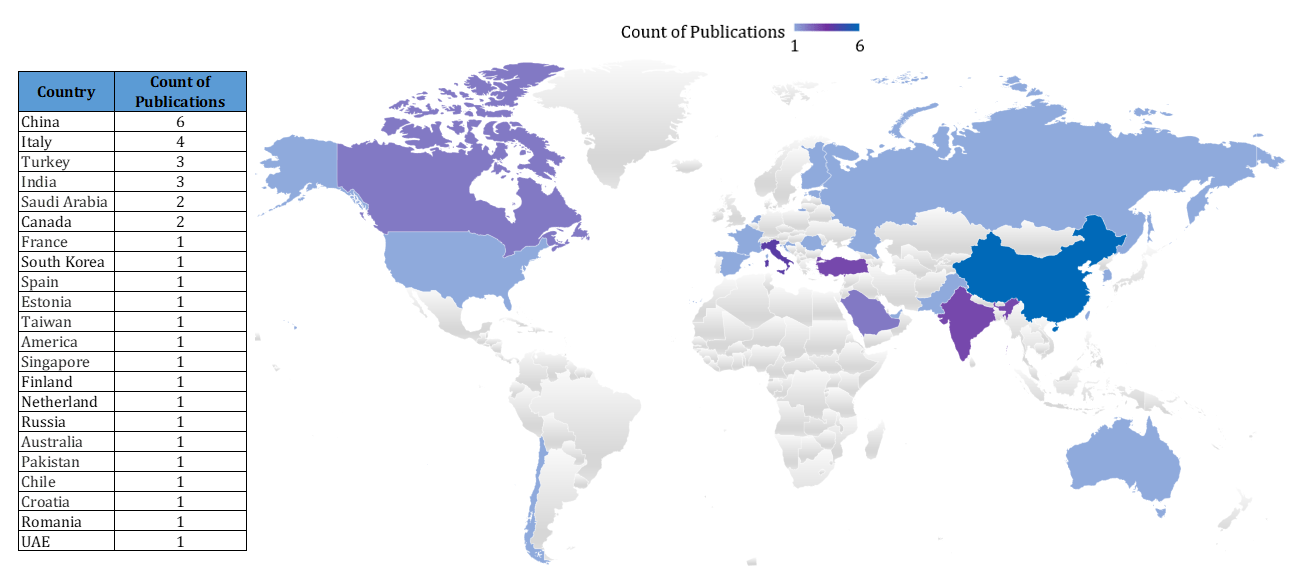}
\caption{Investigating breast cancer diagnosis with XAI: Analyzing first author affiliation belonging to different countries.}\label{fig11}
\end{figure}

\subsection{Criteria For Inclusion and Exclusion}\label{inclusion}
Setting precise inclusion and exclusion criteria guarantees that only pertinent articles on breast cancer diagnosis using XAI are included in this study.
We focused on studies published in reputable journals associated with XAI techniques in breast cancer diagnosis from January 2019 to November 2023. Initially, we found 69 articles using specific keywords (XAI, breast cancer) related to XAI in breast cancer detection and classification. After a thorough review, we excluded articles that did not directly address this topic, duplicates, those lacking full-text availability, or those not aligned with our research. The temporal analysis of research reveals distinct patterns in annual publications, indicative of evolving scholarly interest and engagement with the subject matter. The minimal publication activity in 2019 was followed by a significant increase in research from 2020 to 2021 and a slight decrease in 2022, with a resurgence in 2023 as shown in Figure~\ref{fig12}, 
suggests that factors such as funding, academic priorities, and technological advancements might have influenced these trends. Thus, highlighting the growing importance and potential for further exploration and innovation in integrating XAI into breast cancer diagnosis.

The geographic distribution of research publications reflects the global commitment to advancing the field of breast cancer detection and classification using XAI. The map illustrated in Figure~\ref{fig11}, highlights the number of publications from each country based on author affiliation in this domain. China is at the forefront with six publications, followed by Italy with four and Turkey with three publications, respectively. This widespread participation from various countries highlights the global collaboration and concerted efforts to push forward research in this vital healthcare sector.

\subsection{Search Strategy}\label{search strategy}
To conduct this survey, our search strategy aimed to gather relevant research articles addressing the application of XAI in breast cancer detection. Both automatic and manual search methods were employed to ensure comprehensive coverage of the available literature.
The automatic search was conducted using reputable databases known for hosting scholarly works related to XAI and healthcare, including IEEE Transactions, Arxiv, Google Scholar, Scopus—Elsevier, and Springer. These databases were selected for their extensive collection of relevant research articles in the field.
While other sources such as magazines, working papers, newspapers, books, and blogs contain valuable information, they were not included in this survey due to the lack of rigorous peer-review processes, which could potentially compromise the reliability and quality of the findings.
A variety of general keywords that explored various facets of XAI in breast cancer diagnosis were used for the search, which were derived from the research questions and study titles. These keywords aim to capture a wide range of relevant articles, ensuring a comprehensive understanding of the topic.
Following the retrieval of primary data using these search strings, the analysis phase commenced, focusing on assessing the relevance of the identified research papers to the survey's objectives and predefined inclusion and exclusion criteria.

\subsection{Quality Assessment and Data Extraction}\label{quality assess}
A total of 36 studies were assessed for quality and relevance, and the metadata from these studies was systematically cataloged, ensuring comprehensive documentation of initial findings as outlined by~\cite{kitchenham2010systematic} guidelines for systematic reviews.
Table~\ref{tab:tab3} summarizes the metadata fields extracted from the research studies under investigation. Meanwhile, Table~\ref{tab:tab2} presents the distribution of papers across different quartiles and publishing platforms related to breast cancer detection and classification using XAI. It categorizes publications into Quartile 1 (Q1) and Quartile 2 (Q2) according to Scimago rankings. The Table also includes insights from papers published on platforms like Arxiv and BioArxiv, showcasing the scope of research across diverse publishing venues.
The citation data, obtained from Google Scholar, provided a detailed view of the scholarly impact of the included research articles. This combination of quartile categorization by Scimago Journal Rank (SJR) and citation analysis from Google Scholar offers valuable perspectives on the research's distribution and influence, aiding in a thorough understanding and evaluation of the field within the academic community.

\begin{landscape}
\begin{table}[ht!]
\caption{Publications and citations of XAI methods in the breast cancer domain, ordered by their year of publication (Last accessed on May 2024)}
\label{tab:tab3}
{%
\begin{tabular}{| p{0.6cm}|p{18.8cm}|p{2.5cm}|p{0.5cm}|p{0.6cm}|}
\hline
\textbf{Year} &
  \textbf{Paper Title} &
  \textbf{Authors} &
  \textbf{Cit.} &
  \textbf{Ref.} \\ \hline
2019 &
  Explainable artificial intelligence for breast cancer: A visual case-based reasoning approach\textsuperscript{Q1} &
  Lamy et al. &
  315 &
  ~\cite{lamy2019hierarchical} \\ \hline
2020 &
  Explainable AI (XAI) for anatomic pathology\textsuperscript{Q1} &
  Tosun   et al. &
  68 &
  ~\cite{tosun2020explainable} \\ \hline
2020 &
  A case-based ensemble learning system for explainable breast cancer recurrence prediction\textsuperscript{Q1} &
  Gu   et al. &
  64 &
  ~\cite{gu2020case} \\ \hline
2020 &
  Convolutional neural network-based models for diagnosis of breast cancer\textsuperscript{Q1} &
  Masud   et al. &
  119 &
  ~\cite{masud2020convolutional} \\ \hline
2020 &
  Automated BC detection in digital mammographs of various densities via deep learning\textsuperscript{Q2} &
  Suh   et al. &
  102 &
  ~\cite{suh2020automated} \\ \hline
2021 &
  XAI and susceptibility to adversarial attacks: a case study in the classification of breast ultrasound images\textsuperscript{Arxiv} &
  Rasaee   et al. &
  12 &
  ~\cite{rasaee2021explainable} \\ \hline
2021 &
  Explainable autoencoder-based representation learning for gene expression data\textsuperscript{BioArxiv} &
  Yu   et al. &
  2 &
  ~\cite{yu2021explainable} \\ \hline
2021 &
  Explaining decisions of graph CNN: patient-specific mol. subnetworks responsible for metastasis prediction in BC\textsuperscript{Q1} &
  Cherada et al. &
  81 &
  ~\cite{chereda2021explaining} \\ \hline
2021 &
  ML Approaches to Classify Primary and Metastatic Cancers Using Tissue of Origin-Based DNA Methylation Profiles\textsuperscript{Q1} &
  Modhukur et al. &
  19 &
  ~\cite{modhukur2021machine} \\ \hline
2021 &
  Axillary lymph node metastasis status prediction of early-stage breast cancer using CNN\textsuperscript{Q1} &
  Lee et al. &
  57 &
  ~\cite{lee2021axillary} \\ \hline
2021 &
  One step further into Blackbox: a study of how to build confidence around an AI-based dec. sys. of Br nodule ass.in 2D US\textsuperscript{Q1} &
  Dong et al. &
  28 &
  ~\cite{dong2021one} \\ \hline
2021 &
  XAI Reveals Novel Insight into Tumor Microenvironment Conditions Linked with Better Prognosis in Patients with BC\textsuperscript{Q1} &
  Chakrboty et al. &
  27 &
  ~\cite{chakraborty2021explainable} \\ \hline
2021 &
  Automated scoring of CerbB2/HER2 receptors using histogram-based analysis of immunohistochemistry BC tissue images\textsuperscript{Q1} &
  Kabakcci et al. &
  11 &
  ~\cite{kabakcci2021automated} \\ \hline
2021 &
  Morphological and molecular breast cancer profiling through explainable machine learning\textsuperscript{Q1} &
  Binder et al. &
  110 &
  ~\cite{binder2016layer} \\ \hline
2021 &
  A deep learning classifier for digital breast tomosynthesis\textsuperscript{Q1} &
  Ricciardi et al. &
  25 &
  ~\cite{ricciardi2021deep}\\ \hline
2021 &
  MGBN: Convolutional neural networks for automated benign and malignant breast masses classification\textsuperscript{Q1} &
  Lou et al. &
  13 &
  ~\cite{lou2021mgbn} \\ \hline
2021 &
  A roadmap towards breast cancer therapies supported by explainable artificial intelligence \textsuperscript{Q2} &
  Amorose et al. &
  39 &
  ~\cite{amoroso2021roadmap} \\ \hline
2021 &
  Comparison of feature importance measures as explanations for classification models\textsuperscript{Q2} &
  Saarela et al. &
  221 &
  ~\cite{saarela2021comparison} \\ \hline
2021 &
  Explainable ML can outperform Cox regression predictions and provide insights into breast cancer survival\textsuperscript{Q2} &
  Moncada et al. &
  184 &
~\cite{moncada2021explainable} \\ \hline
2021 &
  Cancer Cell Profiling Using Image Moments and Neural Networks with Model Agnostic Explainability: A Case Study of Breast Cancer Histopathological (BreakHis) Database\textsuperscript{Q2} &
  Kaplun et al. &
  10 &
  ~\cite{kaplun2021cancer} \\ \hline
2022 &
  Benchmarking Counterfactual Algorithms for XAI: From White Box to Black Box\textsuperscript{Arxiv} &
  Moreira et al. &
  6 &
  ~\cite{moreirabenchmarking} \\ \hline
2022 &
  Explainable machine learning of the breast cancer staging for designing smart biomarker sensors\textsuperscript{Q1} &
  Idrees et al. &
  12 &
  ~\cite{idrees2022explainable} \\ \hline
2022 &
  Potential of Non-Contrast-Enhanced Chest CT Radiomics to Distinguish Mol. Subtypes of BC: A Retrospective Study\textsuperscript{Q2} &
  Wang et al. &
  8 &
  ~\cite{wang2022potential} \\ \hline
2022 &
  Applications of Explainable Artificial Intelligence in Diagnosis and Surgery\textsuperscript{Q2} &
  Zhang et al. &
  170 &
  ~\cite{zhang2022applications} \\ \hline
2022 &
  Shape-Based Breast Lesion Classification Using Digital Tomosynthesis Images: The Role of XAI\textsuperscript{Q2} &
  Hussain et al. &
  22 &
  ~\cite{hussain2022shape} \\ \hline
2022 &
  Explainable Multi-Class Classification Based on Integrative Feature Selection for Breast Cancer Subtyping\textsuperscript{Q2} &
  Meshoul et al. &
  2 &
  ~\cite{meshoul2022explainable} \\ \hline
2023 &
  A Hybrid Algorithm of ML and XAI to Prevent Breast Cancer:  A Strategy to Support Decision Making\textsuperscript{Q1} &
  Silva et al. &
  9 &
  ~\cite{silva2023hybrid} \\ \hline
2023 &
  Explainable quantum clustering method to model medical data\textsuperscript{Q1} &
  Deshmukh et al. &
  8 &
  ~\cite{deshmukh2023explainable} \\ \hline
2023 &
  Analyzing breast cancer invasive disease event classification through explainable artificial intelligence\textsuperscript{Q1} &
  Massafra et al. &
  10 &
  ~\cite{massafra2023analyzing} \\ \hline
2023 &
  Information bottleneck-based interpretable multitask network for breast cancer classification and segmentation\textsuperscript{Q1} &
  Wang et al. &
  29 &
  ~\cite{wang2023information} \\ \hline
2023 &
  Applying Ex. ML Models for Detection of BC Lymph Node Metastasis in Patients Eligible for Neoadjuvant Treatment\textsuperscript{Q1} &
  Vrdoljak et al. &
  11 &
~\cite{vrdoljak2023applying} \\ \hline
2023 &
  Peripheral blood mononuclear cell-derived biomarker detection using eXplainable Artificial Intelligence (XAI) provides the better diagnosis of breast cancer\textsuperscript{Q2} &
  Kumar et al. &
  7 &
  ~\cite{kumar2023peripheral} \\ \hline
2023 &
  Applying Deep Learning Methods for Mammography Analysis and Breast Cancer Detection\textsuperscript{Q2} &
  Prodan et al. &
  13 &
  ~\cite{prodan2023applying} \\ \hline
2023 &
  An Explainable Artificial Intelligence Model for the Classification of Breast Cancer\textsuperscript{Q1} &
  Khater et al. &
  5 &
  ~\cite{khater2023explainable} \\ \hline
2023 &
  Cancer Metastasis Prediction and Genomic Biomarker Identification through Machine Learning and eXplainable Artificial Intelligence in Breast Cancer Research\textsuperscript{Q2} &
  Yagin et al. &
  4 &
  ~\cite{yagin2023cancer} \\ \hline
2023 &
  Attention guided Grad-CAM: an improved explainable artificial intelligence model for infrared breast cancer detection\textsuperscript{Q1} &
  Raghvan et al. &
  1 &
  ~\cite{raghavan2023attention} \\ \hline
\end{tabular}%
}
\end{table}
\end{landscape}

\begin{table}[ht!]
\centering
\caption{The Quartile information and number of citations in breast cancer domain.}
\label{tab:tab2}
{%
\begin{tabular}{p{1cm}p{1cm}p{1cm}}
\hline
\multicolumn{1}{c}{\textbf{Quartile}} & \multicolumn{1}{c}{\textbf{Number of Articles}} & \multicolumn{1}{c}{\textbf{Total Citations}} \\ \hline
Q1       & 21 & 885 \\ 
Q2       & 12 & 639 \\ 
Arxiv    & 2  & 14  \\ 
BioArxiv & 1  & 2   \\ \hline
\end{tabular}%
}
\end{table}

\section{Dataset and Modalities for Breast Cancer}\label{Datasets}
Early detection and accurate classification of breast tumors as benign or malignant are crucial in preventing the progression of breast cancer and reducing associated mortality rates. The correct diagnosis enables timely and effective treatment, significantly improving patient outcomes. Various imaging modalities play essential roles in these early detection efforts, with each technology chosen based on specific diagnostic needs.

\begin{figure}[ht!]
\centering
\includegraphics[scale=0.5]{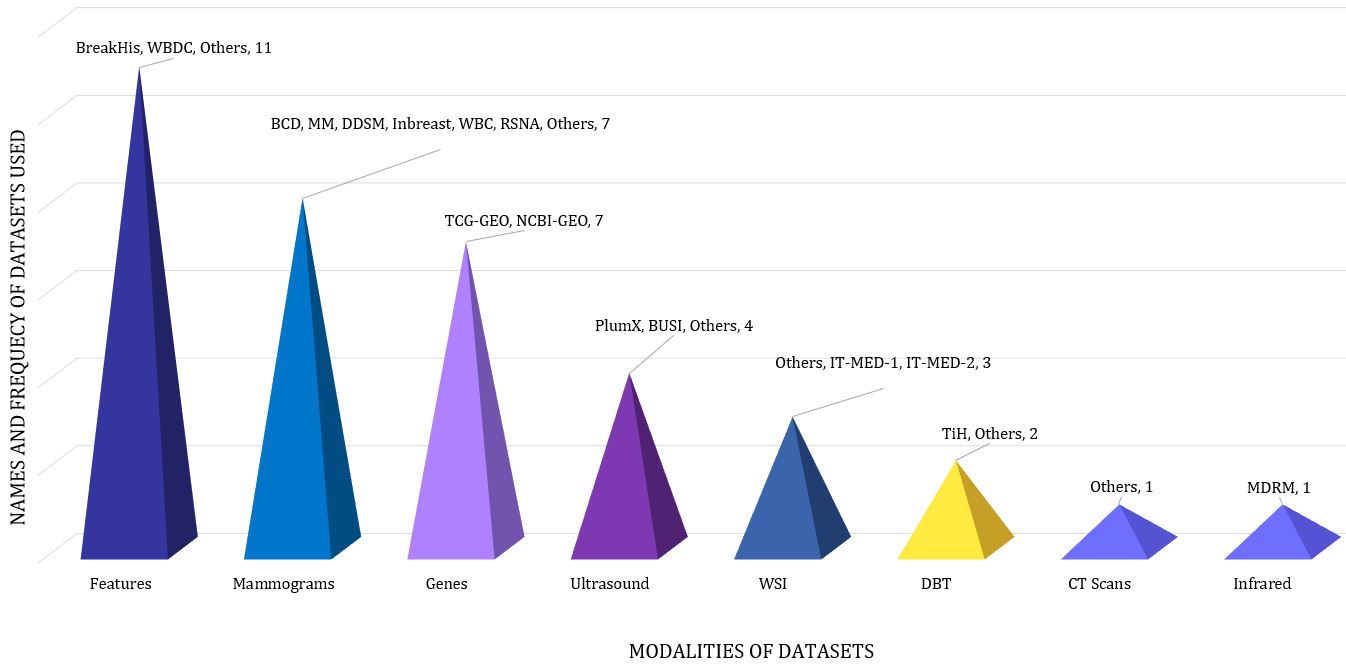}
\caption{Exploring the multi-modal landscape: Datasets driving breast cancer research forward.\\
\scriptsize Features \{BreakHis~\cite{spanhol2015dataset} and others\}, Mammograms \{Breast Cancer Dataset (BCD)~\cite{misc_breast_cancer_14}, Mammographic Mass (MM)~\cite{mammographic_mass_dataset}, Digital Database for Screening Mammography (DDSM)~\cite{lee2017curated}, INbreast~\cite{huang2020dataset}, Wisconsin Breast Cancer (WBC)~\cite{uci_wdbc_1992}, and Radiological Society of North America (RSNA)~\cite{halling2020optimam} and others\}, Genes \{The Cancer Genome Atlas (TCGA) and Gene Expression Omnibus (GEO) (TCG-GEO)~\cite{tcga_website, geo_website}, and National Center for Biotechnology Information (NCBI) and Gene Expression Omnibus (NCBI-GEO)~\cite{labreche2011integrating, piccolo2016integrative}\}, Ultrasound \{Breast Ultrasound Image (PlumX)~\cite{al2020dataset}, Breast Ultrasound Image (BUSI)~\cite{paulo2017breast}, and others\}, Whole Slide Images (WSI) \{ITU-MED-1, ITU-MED-2~\cite{kabakcci2021automated}, and others\}, Digital Breast Tomosynthesis (DBT) \{Two in House (TiH)~\cite{ricciardi2021deep} and others\}, Computed Tomography (CT)\{others\}, and Infrared \{Mastology DRM (MDRM)~\cite{silva2014new}\}. \textbf{Others: Proprietary datasets.}}
\label{fig10}
\end{figure}

\begin{figure}[ht!]
\centering
\includegraphics[scale=0.6]{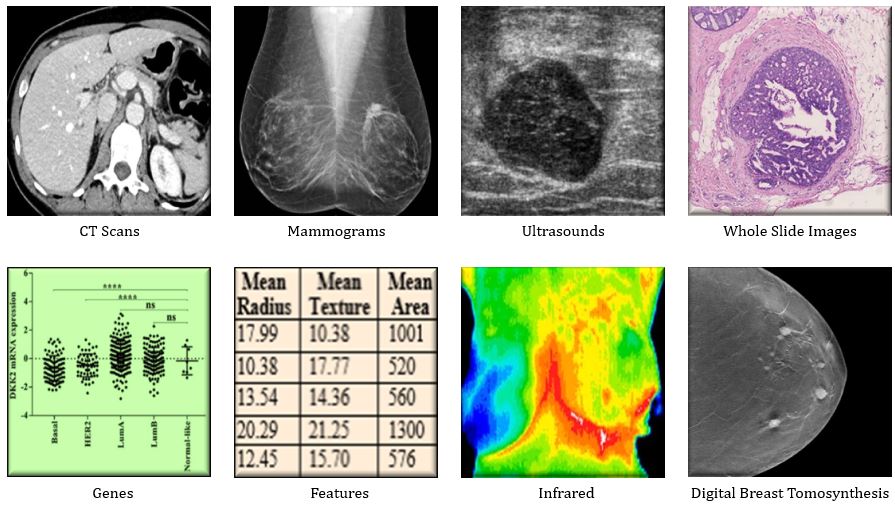}
\caption{Visual Representation of the modalities used in the reviewed studies}
\label{fig90}
\end{figure}

Among the most commonly employed imaging modalities for detecting and diagnosing breast cancer are mammograms, CT Scans, ultrasounds, and WSI. These modalities offer unique benefits that are instrumental in identifying cancerous changes at an early stage. For instance, mammograms are a primary tool in breast cancer screening programs due to their effectiveness in detecting early signs of cancer through low-dose X-ray images. These images are easy to capture and are often used as the first test in breast cancer detection. In addition to traditional imaging techniques, integrating advanced modalities like DBT and various genomic and histopathological datasets enhances the depth and accuracy of breast cancer analysis. The visual representation of each modality is depicted in Figure~\ref{fig90}.
The literature review has indicated that XAI methods have been significantly employed across various modalities, with the features dataset emerging as a prominently utilized resource for enhancing the transparency and interpretability of models in breast cancer research, as exhibited in Figure~\ref{fig10}. A feature dataset in breast cancer research encompasses specific attributes gathered from various diagnostic tools, including imaging, genetic tests, and biochemical analysis. Imaging features from mammograms detail tumor characteristics like shape and texture~\cite{binder2021morphological}. Histopathological features involve the microscopic analysis of tissue samples~\cite{kaplun2021cancer}, and genomic features highlight gene expression and mutations~\cite{chakraborty2021explainable}, ~\cite{meshoul2022explainable},~\cite{yu2021explainable}, ~\cite{yagin2023cancer}. Biochemical features such as levels of enzymes, including lactate dehydrogenase, alkaline phosphatase, and aromatase, as well as hormone assays, provide insights into biological activities associated with cancer~\cite{moncada2021explainable}, ~\cite{massafra2023analyzing}. Clinical features integrate personal data such as age and family history\cite{vrdoljak2023applying},~\cite{silva2023hybrid}. 

These comprehensive modalities encompass not just the screening and detection but also the segmentation and classification of breast cancer, utilizing a broad range of datasets and imaging technologies. Each modality contributes uniquely to the complex process of cancer diagnosis, from the detailed anatomical insights provided by the genetic information revealed through genomic databases. This multi-modal strategy is critical for developing a nuanced understanding of breast cancer, leading to more personalized and effective treatment strategies. Thus, the blend of traditional and advanced imaging modalities, enriched by diverse datasets, highlights the multidisciplinary nature of contemporary breast cancer research and treatment. 

\section{Explainable Artificial Intelligence (XAI)}\label{XAI}
\begin{figure}[!htbp]
\centering
\includegraphics[width=14cm, height=3.5cm]{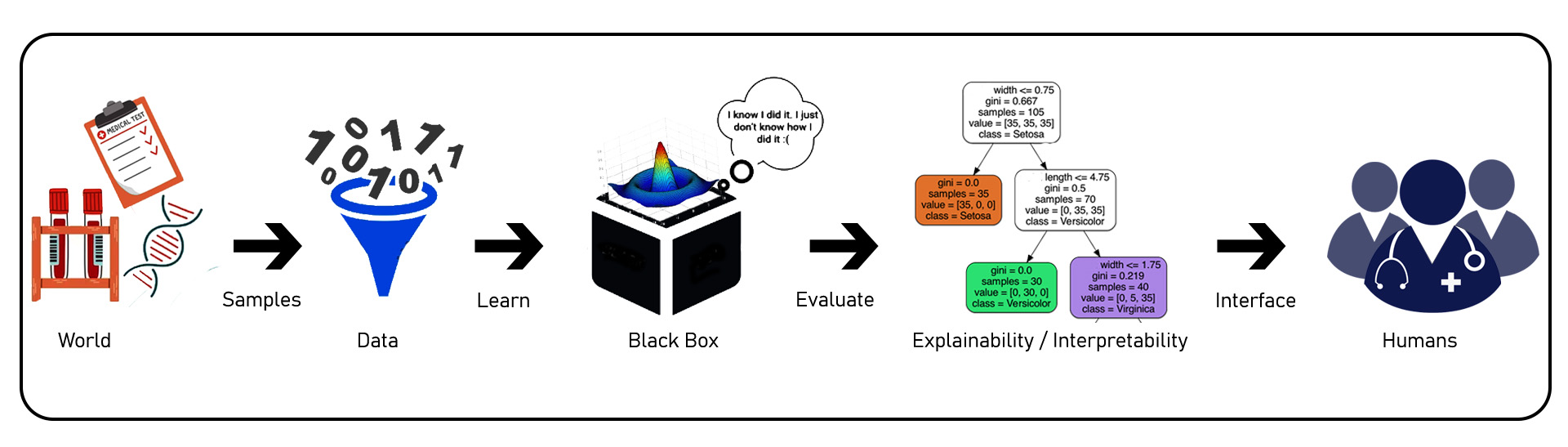}
\caption{Concept of black box models focusing the essential aspect of explainability in machine learning~\cite{1044_Khan}}
\label{fig15}
\end{figure}

Recently, the ML/DL models have been extensively used in the healthcare domain to predict and diagnose different diseases, such as Autism Spectrum Disorder (ASD)~\cite{sharif2022novel}, and different kinds of cancer, for instance, lung cancer~\cite{li2022machine}, breast cancer~\cite{ShahKAS22}, etc. These models are inherently black boxes and lack the interpretability of the predicted decisions. It can eventually lead to distrust in the model and ultimately limit its clinical utility. With the advent of XAI, researchers are focusing on integrating the ML/DL models with XAI techniques to better understand the decisions by providing transparency on how these models make predictions, as explained in Figure~\ref{fig15}.

The machine-learning pipeline can be improved by initiating the concept of explainability~\cite{1044_Khan}. In this article, we focus on how XAI investigates the impact of specific features on the models' predictions and their corresponding insights in the detection and classification of primary and secondary breast cancer. 
Breast cancer classification and detection typically involve analyzing medical images, such as mammograms, using deep learning algorithms. XAI techniques are useful in overcoming the opacity of black box models.

This research is a comprehensive examination of individual studies, aiming to pinpoint the algorithms and methodologies proposed. Additionally, it facilitates the categorization of studies according to the machine learning techniques utilized in XAI. This approach showcases the relationship between ML/DL models and XAI techniques, as outlined in Table~\ref{tabt:tab-1}.

\begin{figure}[ht!]
\includegraphics[width=18cm, height=21.5cm]{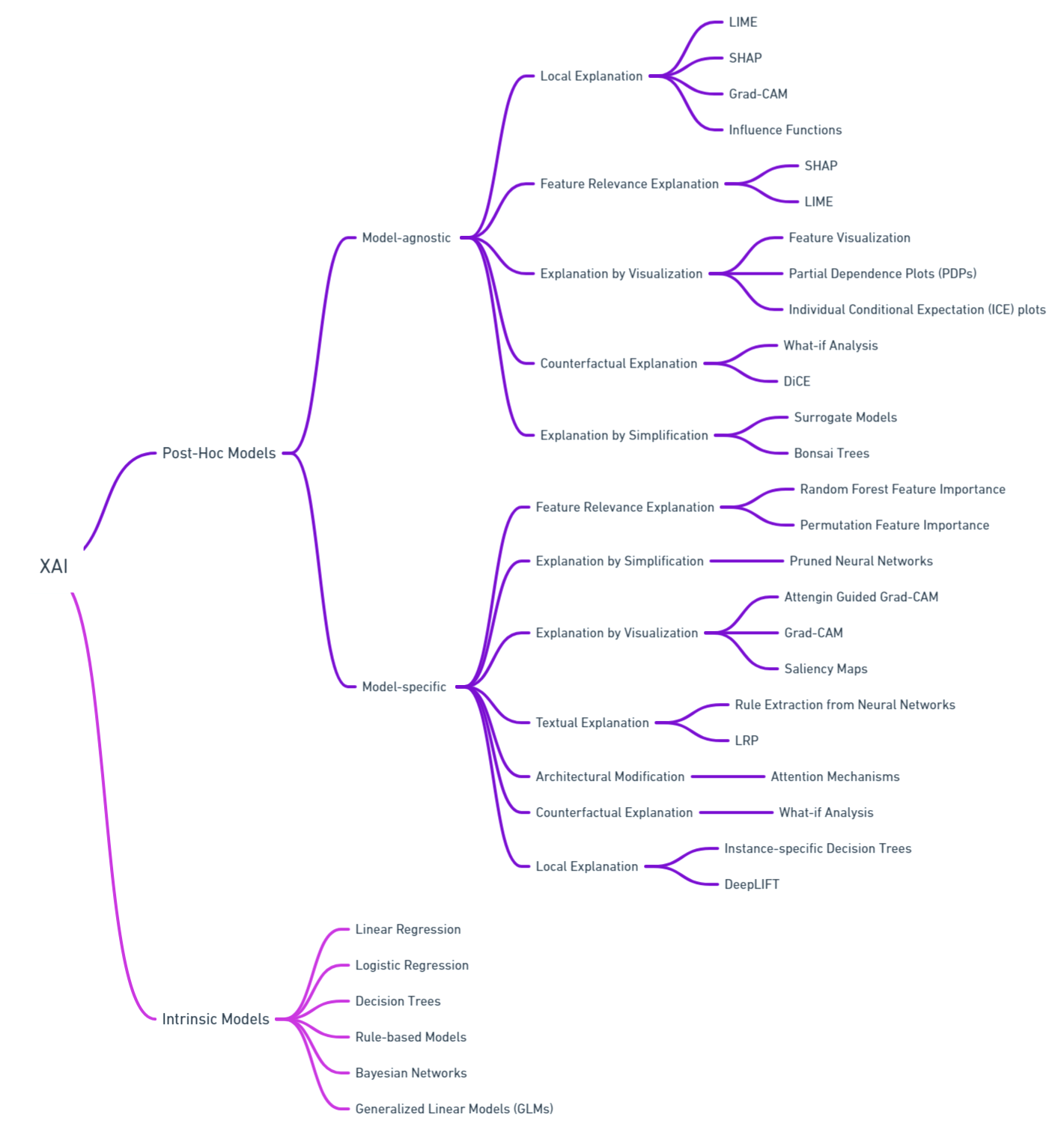}
\caption{Comprehensive Taxonomy of Explainable Artificial Intelligence (XAI) models}\label{fig14}
\end{figure}

\begin{table}[ht!]
\centering
\caption{Correlation Between Machine Learning/Deep Learning Models and Explainable AI (XAI) Techniques}
\label{tabt:tab-1}
	
	\begin{tabular}{|p{3cm}|p{1cm}|p{1cm}|p{1.2cm}|p{0.8cm}|p{1cm}|p{0.8cm}|p{1cm}|p{0.8cm}|p{0.7cm}|p{0.9cm}|p{0.8cm}|}
\hline
 &
  \textbf{SHAP     ~\cite{lundberg2017}} &
  \textbf{LIME          ~\cite{ribeiro2016}} &
 \textbf{GRAD-CAM      ~\cite{selvaraju2017grad}} &
 \textbf{CAM       ~\cite{zhou2016learning}} &
  \textbf{AGG-CAM      ~\cite{yagin2023cancer}} &
 \textbf{CBR     ~\cite{althoff2001case}} &
  \textbf{GLRP       ~\cite{chereda2021explaining}} &
  \textbf{ADR         ~\cite{amoroso2021roadmap}} &
  \textbf{LRP     ~\cite{bach2015pixel}} &
 \textbf{PIMP   ~\cite{altmann2010permutation}} &
  \textbf{PDP    ~\cite{gianfagna2021explainable}} \\ \hline
RF        ~\cite{breiman2001random}          & $\surd$ & $\surd$&      &      &      &      &      &      &      & $\surd$& $\surd$\\ \hline
RF (Survival)~\cite{ishwaran2008random} & $\surd$ &      &      &      &      &      &      &      &      &      &      \\ \hline
Gradient Boost~\cite{friedman2001greedy}           & $\surd$ &      &      &      &      &      &      &      &      &      &      \\ \hline
AdaBoost~\cite{freund1997decision}               & $\surd$ &      &      &      &      &      &      &      &      &      &      \\ \hline
GoogLeNet~\cite{szegedy2015going}                &      &      & $\surd$ &      &      &      &      &      &      &      &      \\ \hline
AlexNet~\cite{krizhevsky2012imagenet}                  &      &      & $\surd$ &      &      &      &      &      &      &      &      \\ \hline
VGG~\cite{simonyan2014very}                      &      &      & $\surd$ &      & $\surd$ &      &      &      &      &      &      \\ \hline
Xception~\cite{chollet2017xception}                 &      &      & $\surd$&      &      &      &      &      &      &      &      \\ \hline
ResNet~\cite{he2016deep}                   &      & $\surd$ & $\surd$ & $\surd$ &      &      &      &      &      &      &      \\ \hline
DenseNet~\cite{huang2017densely}                 &      & $\surd$ &$\surd$&      & $\surd$&      &      &      &      &      &      \\ \hline
Mask R-CNN~\cite{he2017mask}               &      & $\surd$ &      &      &      &      &      &      &      &      &      \\ \hline
Graph CNN~\cite{defferrard2016convolutional}                &      &      &      &      &      &      & $\surd$ &      &      &      &      \\ \hline
MobileNet~\cite{howard2017mobilenets}                &      & $\surd$ & $\surd$ &      &      &      &      &      &      &      &      \\ \hline
SqueezeNet~\cite{iandola2016squeezenet}               &      & $\surd$ & $\surd$&      &      &      &      &      &      &      &      \\ \hline
ResNeXt~\cite{xie2017aggregated}                  &      & $\surd$ & $\surd$ &      &      &      &      &      &      &      &      \\ \hline
MIB-Net~\cite{wang2023information}                  &      &      & $\surd$ &      &      &      &      &      &      &      &      \\ \hline
DarkNet~\cite{redmon2017yolo9000}                  &      &      & $\surd$ &      &      &      &      &      &      &      &      \\ \hline
EfficientNet~\cite{koonce2021efficientnet}             &      &      & $\surd$ &$\surd$&      &      &      &      &      &      &      \\ \hline
XGBoost~\cite{chen2016xgboost}                 & $\surd$ & $\surd$ &      &      & $\surd$ & $\surd$ &      &      &      &      &  $\surd$ \\ \hline
SVM~\cite{boser1992training}                      & $\surd$ & $\surd$ &      &      &      &      &      &      &      &      &      \\ \hline
LightGBM~\cite{ke2017lightgbm}                & $\surd$ &      &      &      &      &      &      &      &      &      &      \\ \hline
CatBoost~\cite{dorogush2018catboost}                & $\surd$ &      &      &      &      &      &      &      &      &      &      \\ \hline
Autoencoder~\cite{hinton2006reducing}             & $\surd$ &      &      &      &      &      &      & $\surd$ &      &      &      \\ \hline
Clustering~\cite{hansen1997cluster}               &      & $\surd$ &      &      &      &      &      &      &      &      &      \\ \hline
MaXViT~\cite{tu2022maxvit}                   &      &      &      & $\surd$ &      &      &      &      &      &      &      \\ \hline
Naïve Bayes~\cite{langley1992analysis}              & $\surd$ & $\surd$ &      &      &      &      &      &      &      &      &      \\ \hline
Bag-of-Words~\cite{manning2009introduction}             &      &      &      &      &      &      &      &      & $\surd$ &      &      \\ \hline
LSTM~\cite{hochreiter1997long}                     &      &      &      &      &      &      &      &      &      &      &      \\ \hline
\end{tabular}%
\end{table}

\section{An Insight into XAI Techniques}\label{XAI_techs}
A comprehensive taxonomy of XAI techniques is presented in Figure~\ref{fig14}. In general, XAI models can be categorized into two types, including \textbf{intrinsic} and \textbf{post-hoc models}. It is possible to explain and comprehend intrinsic models, also called transparent models. Typical examples of transparent models include linear regression, logistic regression, and decision trees~\cite {arrieta2020explainable}. On the other hand, post-hoc is categorized into \textbf{model-specific} and \textbf{model-agnostic}.


XAI explanations can also be categorized into two main types based on their scope, which includes \textbf{global} and \textbf{local}. A global explanation offers a comprehensive overview of the entire model, while a local explanation focuses on a single prediction or output. The specifics of various XAI models are outlined in the following subsections, based on their prevalence in ML/DL applications within the breast cancer domain.
The next section provides a detailed discussion of widely used XAI models and reviews corresponding research works in the context of breast cancer.

\subsection{\underline{SH}apley \underline{A}dditive  ex\underline{P}lanations (SHAP)}\label{shap}
Lundberg \cite{lundberg2017} presented a unified framework, SHAP for predicting interpretations. Each feature is assigned an important value for a particular prediction. 

SHAP is a post-hoc and model-agnostic, explainable model in machine learning. It can explain the output of any black box model by computing the contribution of each feature to the prediction. SHAP values provide a measure for the impact of each feature in a given data instance on the model's output, offering both a global and local understanding of the model.

The mathematical foundation of SHAP is rooted in game theory ~\cite{young1985monotonic}, specifically in the concept of Shapley values. Its design ensures equitable distribution of the "payout" (prediction) among the "players" (features).\\
\textbf{SHAP Equation:}
The primary SHAP equation is formulated as follows:
\begin{equation}\label{eq:1}
    \phi(f, x) = \sum_{z' \subseteq x'} \frac{|z'|!\,(M - |z'| - 1)!}{M!} \left[f_x(z') - f_x(z' \setminus i)\right]
\end{equation}

Where, $\phi(f, x)$ represents the SHAP value for a prediction $f$ given the input $x$, $z'$ is a subset of the features, $M$ is the total number of features, $|z'|$ is the number of non-zero entries in $z'$ and $f_x(z')$ is the prediction for the subset $z'$, and $f_x(z' \setminus i)$ is the prediction without a particular feature.

For the given equation \ref{eq:1} the underlying properties taken into consideration are as follows;

\begin{enumerate}

\item \textbf{Local Accuracy:} 
   This ensures that the explanation model's output $g(x')$ matches the output of the original model $f(x)$ for any input $x$. The relationship is given by:
   \begin{equation}
    f(x) = g(x') = \phi_0 + \sum_{i=1}^{M} \phi_i x'_i 
    \end{equation}
   \begin{itemize}
       \item $\phi_0$ is the base value (model's output for the average of all inputs).
       \item $\phi_i$ are the SHAP values attributing the difference from the base value to each feature.
   \end{itemize}

\item \textbf{Missingness:}
   This principle asserts that if a feature is absent (or has a value of zero) in the input, it should not influence the prediction. It's mathematically represented as:
   \begin{equation}\label{eq:2}
     x'_i = 0 \Rightarrow \phi_i = 0
 \end{equation}
   This ensures that features not present do not impact the model's output.

\item \textbf{Consistency:}
   The SHAP values should remain consistent when the influence of a feature on the model output increases or remains unchanged. If increasing or keeping a feature constant increases the output, the SHAP value for that feature should not decrease. This is crucial to ensuring that more influential features are always given higher importance.
  
   Let $f_x(z') = f(h_x(z'))$ and $z'_i \text{ denote setting } z'_i = 0$. For any two models, $f$ and $f'$, if 
   
\begin{equation}\label{eq:3}
    f_x'(z') - f_x'(z' \setminus i) \geq f_x(z') - f_x(z' \setminus i)
\end{equation}

\begin{equation} \forall z' \in \{0,1\} ^M, \text{then } \phi_i(f', x) \geq \phi_i(f, x) \end{equation} 

SHAP values provide a mathematically rigorous, fair, and consistent method for interpreting machine learning model predictions, making them an invaluable tool in explainable AI.

\end{enumerate}

\subsubsection{SHAP: A Post-hoc Explanation Method in Breast Cancer Diagnosis}
Many breast cancer detection and classification techniques make use of SHAP to explain the results to doctors. SHAP can be used with \textbf{feature-based datasets} to pinpoint the pertinent variables and assess their influence on the prediction's significance. In this connection, some researchers have worked on the survival analysis of breast cancer combined with XAI techniques. 
For instance, Moncada et al.~\cite{moncada2021explainable} designed a system using the Cox Proportional Hazards (CPH) model against three other ML models (i.e., Random Survival Forest (RSF), Survival Support Vector Machine (SSVM), and Extreme Gradient Boosting (XGBoost)) for the survival analysis of primary invasive non-metastatic breast cancer patients. The researchers opted to exclusively include female patients in the Netherlands diagnosed with primary invasive non-metastatic breast cancer between 2005 and 2008 and who underwent curative surgery, which could either be breast-conserving surgery or mastectomy. The authors evaluated the models' performance using the c-index. They selected nine features for comparison, encompassing age, tumor characteristics, (hormonal) receptor statuses, clinical and pathological TNM (Tumour, Node, Metastasis) staging, and the presence of removed and positive lymph nodes. The SHAP model is employed to provide explainability for the results of the reference model (CPH) and other machine learning models. Among these models, XGBoost is the best performer, achieving a c-index of approximately 0.73. Moreover, SHAP values explore the influence of specific features on model predictions, a task that can be intricate even for experts.
Zhang et al. ~\cite{zhang2022applications} surveyed recent trends in surgical applications and medical diagnosis using XAI based on findings across various research platforms, described various XAI-enabled methods for medical XAI applications, and finally discussed the achieved challenges and research directions. Zhang et al. also propose a case study for accurate breast cancer predictions and diagnosis. 
The authors employed a real-world dataset on breast cancer from Wisconsin (Diagnostic), comprising 569 patients. The dataset includes numerical features extracted from digitized images of fine needle aspirates (FNA) of breast masses. It contains 357 benign and 212 malignant as per class distribution.
The authors applied the intrinsic XAI method i.e. Rule-based for the breast cancer classification task. The proposed rule-based model attained an accuracy of 60.81\%, precision of 60.95\%, recall of 99.04\%, and F1-score of 75.46\%. Additionally, the authors applied some post-hoc XAI methods including SHAP, LIME (as discussed in Section \ref{lime}), and PDP (as discussed in Section \ref{pdp}) for a black-box model Random Forest. It was identified that Random Forest outperformed the rule-based approach, achieving an accuracy of 95.91\%, precision of 97.09\%, recall of 96.15\%, and an F1-score of 96.62\%. The post-hoc XAI method SHAP aided in interpreting the results of the black box model.SHAP identified that the features including "worst parameter", "worst area", "worst radius", "worst concave points", "worst texture" and "mean texture" have a positive effect on predictions.
Likewise, Massafra et al.\cite{massafra2023analyzing} investigated invasive disease events (IDE) associated with breast cancer in 486 patients registered at IRCCS Istituto Tumori "Giovanni Paolo II" in Bari, Italy. They employed SHAP to provide a cohesive explanation for the classifier's predictions. The authors employed various classifiers, including RF, SVM, XGBoost, and Naive Bayes (NB). Consistent with findings from Moncada et al.~\cite{moncada2021explainable}, it was found that the XGBoost model outperformed the other three models. The XGBoost model achieved median AUC values of 93.7\% and 91.7\% for predicting IDE over 5 and 10 years, respectively. Shapley values were instrumental in identifying the key IDE-driving features across these two time frames. Specifically, they revealed that within the 5 years, \texttt{age}, \texttt{tumor diameter}, \texttt{surgery type}, and \texttt{multiplicity} were predominant. 

On the other hand, Vrdoljak et al.\cite{vrdoljak2023applying} utilized four ML classifiers XGBoost, RF, Logistic Regression, and Univariate Logistic Regression to predict breast cancer lymph node metastases in patients eligible for neoadjuvant systematic therapy (NST). Given the limitations of radiological methods, up to 30\% of patients are misdiagnosed when determining axillary lymph node status, so it is deemed necessary to employ additional assessment methods for lymph node status. For this research, data were collected from all Croatian hospitals, and out of 8381 breast cancer patients, 719 were deemed eligible for NST. The authors evaluated the model's explainability using Shapley values. The results of this research underscore \texttt{patient age}, \texttt{tumor size}, and \texttt{Ki-67} as the three primary indicators of breast cancer risk. Similar to previous approaches including Moncada et al.\cite{moncada2021explainable} and Massafra et al.\cite{massafra2023analyzing}, XGBoost exhibited the best performance in this experiment, achieving a mean AUC of 0.762 (95\% CI: 0.726–0.794).
 
In another work, Silva et al.~\cite{silva2023hybrid} presented a hybrid solution combining ML and XAI algorithms for a clinical decision support system (CDSS) for breast cancer prevention. The authors have used the CRISP-ML (Q) methodology for standard data management. The case study is based on the publicly available dataset ~\cite{nindrea2021dataset} of Indonesian patients. XGBoost algorithm is used for classifying breast cancer patients and is combined with the SHAP model for model explainability. The XGBoost algorithm resulted in an accuracy of 85\%. The interpretability of XGBoost is increased with the help of SHAP values. It identified the two most relevant variables, \texttt{high-fat diet} and \texttt{breastfeeding}, that have a high impact on classifying breast cancer patients. 

To attain a deeper comprehension of the fundamental factors influencing the dataset and its predictive results., some authors have incorporated a diverse range of XAI methods along with SHAP in breast cancer classification and detection use cases. For example, Khater et al.~\cite{khater2023explainable} employed three different model-agnostic methods, including SHAP, permutation importance (PIMP) (as discussed in Subsection \ref{pimp}), and partial dependence plot (PDP) (as mentioned in Subsection \ref{pdp}), to provide explanations for ML model predictions in the breast cancer classification domain. The research involved training various ML algorithms, including SVM, KNN, RF, and XGBoost, for diagnosing breast cancer over genomic datasets such as Wisconsin Breast Cancer (WBC) and Wisconsin Diagnostic Breast Cancer (WDBC) datasets. The KNN achieved the highest accuracy of 97.7\% and a precision of 98.2\% with the Wisconsin breast cancer dataset, and an accuracy of 98.6\% through the artificial neural network, resulting in a precision of 94.4\% based on the dataset. SHAP indicated the contribution of features to the prediction of the ML models. Shapley values were computed for features representing only extreme values to assess their impact on breast cancer class predictions. 

Some authors have worked with \textbf{genomics datasets} such as The Cancer Genome Atlas (TCGA) and National Center for Biotechnology Information-Gene Expression Omnibus (NCBI-GEO) in the breast cancer domain.
Chakraborty et al.~\cite{chakraborty2021explainable} proposed XAI usage to enhance the explainability of prognosis management and $\geq$ 5-year survival rate predictability in patients suffering breast cancer. The authors used TCGA invasive breast cancer data obtained from the cbioPortal, while estimates of immune cell composition were extracted from bulk RNA sequencing data using TIMER2.0, which employed EPIC, CIBERSORT, TIMER, and xCell computational processes.
The authors employed XGBoost and SHAP techniques for survivability models of breast cancer, focusing on tumor microenvironment (TME) conditions, encompassing immune and tumor-associated cells. The authors performed the global SHAP analysis to determine the hierarchy of importance of TME cells on patients' survival rates. They also conducted a local SHAP analysis to pinpoint the thresholds of TME cells, beyond or below which survival rates may escalate. 
According to the SHAP analysis findings, maintaining a B cell fraction above 0.025 relative to all cells in a sample, ensuring the \texttt{M0 macrophage fraction} is below 0.05 relative to the total immune cell content, and having \texttt{NK T cell} and \texttt{CD8+ T cell fractions} above 0.075 and 0.25, respectively, based on cancer type-specific arbitrary units, could potentially improve the $\geq$ five-year survival rate in breast cancer patients.

Similarly, Yu et al.~\cite{yu2021explainable} developed a tool, XAE4Exp (eXplainable AutoEncoder for Expression data) to facilitate a SHAP-supported explainable AE-based representation learning. 
XE4Exp quantitatively assesses the contributions of each gene to the hidden structure learned by an AE, significantly extending the AE outcomes. This tool is applied to the TCGA breast cancer gene expression data and identifies genes that are not differentially expressed and pathways in various cancer-related classes. 
Additionally, the SHAP-based explanations help to reveal sensible pathways.

Likewise, Meshoul et al.~\cite{meshoul2022explainable}, proposed a multi-stage feature selection (FS) framework for BC subtype classification using "The Cancer Genome Atlas" (TCGA) multi-omics data. The authors used four ML models, including SVM, RF, extra trees, and XGBoost. Additionally, SHAP was used to explain specific features that influenced classification. The utilization of SHAP helped to identify the most significant features for instance \texttt{DNA, mRNA}, and \texttt{CNV}, which aid in the identification of a particular subset that influenced classification.

Furthermore, Kumar et al. ~\cite{kumar2023peripheral} investigated the potential diagnostic biomarkers for breast cancer using XAI on XGBoost ML models trained on a binary classification NCBI-GEO dataset containing the expression data of Peripheral Blood Mononuclear Cells (PBMCs) from 252 breast cancer patients and 194 healthy women over NCBI-GEO dataset. Further, the authors effectively added SHAP values into the XGBoost model, and ten important genes associated with breast cancer development were determined to be effective potential biomarkers. This research identified that \texttt{SVIP, BEND3, MDGA2, LEF1-AS1, PRM1, TEX14, MZB1, TMIGD2, KIT}, and \texttt{FKBP7} are key genes that impact model prediction. These genes may act as early, non-invasive diagnostic and prognostic biomarkers for breast cancer patients.

Yagin et al.~\cite{yagin2023cancer} presented a model combining ML techniques such as LightGBM, CatBoost, XGBoost, GBTBoost, and AdaBoost, along with SHAP for predicting metastatic breast cancer. The study revealed key genomic biomarkers in patients suffering metastasis. Based on the SHAP results, \texttt{TSPYL5} emerged as the most significant gene for predicting breast cancer metastasis.
Recent research studies have uncovered that radiomic features derived from computed tomography (CT) or magnetic resonance imaging (MRI) hold the potential to distinguish between luminal and non-luminal breast cancer (BC). 

Wang et al.\cite{wang2022potential} developed a radiomics model based on chest CT to differentiate between luminal and non-luminal types, achieving an area under the curve (AUC) values of 0.842 in the training set and 0.757 in the test set. This suggests that chest CT radiomics could offer a novel approach to identifying breast cancer molecular subtypes. Additionally, the authors employed SHAP dependence plots to visualize the final model predictions, illustrating how individual features influence the output of the LASSO\_SVM prediction model. SHAP values were utilized to quantify the contribution of each feature to the predicted outcome\cite{rodriguez2019interpretation}.
A comprehensive discussion of the pros and cons of the SHAP technique in breast cancer diagnosis is presented in Section~\ref{discussion}.

\subsection{\underline{C}lass \underline{A}ctivation \underline{M}ap and \underline{G}radient-weighted \underline{C}lass \underline{A}ctivation \underline{M}ap (Grad-CAM)}\label{grad&cam}

{Class Activation Map (CAM) and Gradient-CAM (Grad-CAM)}\label{cam}
Zhou et al., introduced class activation maps (CAM) using Global Average Pooling (GAP) in CNNs after the final convolutional layer to decrease the size of the image and reduce the parameters to avoid overfitting. CAM helps to identify the discriminative regions of an image that affect the output of CNNs by performing global average pooling on the convolutional feature maps and using those as features for a fully connected layer that produces the desired output. Class activation weights are then computed based on the gradients of the predicted class score relying on the feature maps. These weights indicate the importance of each feature map for the target class.\\

\textbf{CAM} generates heatmaps employing \(f_k (x, y)\), the pixel-wise activation of unit \(k\) across spatial coordinates \((x, y)\) in the last convolutional layers, weighted by \(w_k^c\), the coefficient corresponding to unit \(k\) for class \(c\) at pixel \((x, y)\) is formulated as follows:
\begin{equation}\label{eq:4}
M_c(x,y) = \sum_k w_k^c f_k(x,y)
\end{equation}




\textbf{Grad-CAM}\label{gradcam}
Selvaraju et al. \cite{selvaraju2017grad} presented, a technique for generating visual explanations from convolutional neural networks (CNNs) by highlighting important regions in an image for predicting a specific concept, making CNN-based models more interpretable. It is a post-hoc, model-specific because it utilizes the gradients and feature maps specific to a given convolutional neural network to generate visual explanations, highlighting how certain regions of the input contribute to the output; however, it is also model-agnostic in the sense that it can be applied across different CNN architectures without needing modifications specific to each model's architecture~\cite{khan2022model}.

The mathematical foundation of Grad-CAM is based on the gradients of the target class with respect to the feature maps. These gradients indicate the importance of each feature map for the target prediction. Grad-CAM is particularly useful for understanding and debugging CNNs in tasks such as image classification, image captioning, and visual question answering.

\textbf{Grad-CAM Equation:}
The primary Grad-CAM equation is formulated as follows:
\begin{equation}\label{eq:gradcam}
    L_{Grad-CAM}^c = \text{ReLU}\left(\sum_k \alpha_k^c A^k\right)
\end{equation}

Where \( L_{Grad-CAM}^c \) represents the Grad-CAM localization map for class \( c \), \( A^k \) is the \( k-th \) feature map of the convolutional layer and \( \alpha_k^c \) is the importance weight of the \( k \)-th feature map for class \( c \).

For the given equation \ref{eq:gradcam}, the underlying properties taken into consideration are as follows:

\begin{enumerate}

\item \textbf{Localization:}
   Grad-CAM localizes the important regions in the image that contribute to the prediction of a specific class. This is achieved by combining the feature maps \( A^k \) with the importance weights \( \alpha_k^c \) to produce a class-specific localization map.
    The ReLU activation function ensures that only the positive contributions are highlighted in the localization map.
   The ReLU activation function is applied to the weighted sum of the feature maps to ensure that only the positive contributions are included in the localization map. This step helps in highlighting the regions that positively influence the prediction of the target class

\item \textbf{Class Discrimination:}
   The importance weights \( \alpha_k^c \) are computed using the gradients of the target class with respect to the feature maps. This ensures that the localization map is specific to the target class and highlights the regions that are important for predicting that class.

\item \textbf{Global Average Pooling:}
   The importance weights \( \alpha_k^c \) are obtained by performing global average pooling on the gradients of the target class with respect to the feature maps:

   \begin{equation}\label{eq:alpha}
    \alpha_k^c = \frac{1}{Z} \sum_i \sum_j \frac{\partial y^c}{\partial A_{ij}^k}
   \end{equation}
   where \( Z \) is the number of pixels in the feature map, \( y^c \) is the score for class \( c \), and \( \frac{\partial y^c}{\partial A_{ij}^k} \) is the gradient of the score for class \( c \) with respect to the activation \( A_{ij}^k \) at spatial location \( (i, j) \) in the feature map \( A^k \).
   This aggregation step ensures that the importance of each feature map is considered across the entire spatial dimensions of the feature map.

\end{enumerate}

\subsubsection{CAM and Grad-CAM: Post-Hoc Explanation Methods in Breast Cancer Diagnosis}
In ~\cite{prodan2023applying}, the authors utilized CAM for different CNN models for classifying mammography images of breast cancer patients, including ResNet, EfficientNet, and MaxVit. Subsequently, two visualization techniques were employed to enhance comprehension of the outcomes: class activation maps (CAM) and centered bounding boxes. The approach provided a more intuitive insight into the model's decision-making mechanism by spotlighting the particular area of the image most pertinent for classification. The region with the highest activation value is pinpointed, and a centered bounding box is generated around it. 


Grad-CAM is one of the most widely utilized explainable methods in the breast cancer domain, along with radiomics such as ultrasound, mammography, CT scans, and DBT. Also, with some genomic biomarkers.
For instance, Masud et al. ~\cite{masud2020convolutional} proposed a custom model for classifying ultrasound images using eight pre-trained CNN models with a transfer learning mechanism. It is observed that ResNet50 obtained the best accuracy of 92.4\% with the Adam optimizer, and VGG16 achieved the maximum AUC 0.97 score. Furthermore, the authors used Grad-CAM heatmap visualization to investigate how well the proposed model classifies breast cancer. Grad-CAM visualizes a portion of an image through a heatmap of a class label and results in benign or malignant classes based on the prediction probability. It is observed that Grad-CAM output perfectly focuses on the critical areas of images to classify cancers.

Similarly, Dong et al.~\cite{dong2021one}, proposed a classification method for predicting primary breast lesions using DenseNet-121 over 2D breast ultrasound images. The data comprised a total of 785 2D breast ultrasound images acquired from 367 females from April 2017 to July 2019 from Shenzhen People’s Hospital. Additionally, authors have suggested a novel region of evidence (ROE) by incorporating GRAD-CAM to help physicians and patients understand AI decisions about the initial screening of ultrasound images of breast cancer. The results indicate an AUC of 0.899 and 0.869, respectively. The accuracy, sensitivity, and specificity of the model with coarse ROIs are 88.4\%, 87.9\%, and 89.2\%, and with fine ROIs are 86.1\%, 87.9\%, and 83.8\%. While some researchers have worked on \textbf{mammography datasets} such as Suh et al.~\cite{suh2020automated} presented an automated breast cancer detection on digital mammograms using Dense-Net-169 and Efficient-Net-B5 on a private dataset to predict the availability of malignancy of lesions. The CNN models could detect malignant lesions efficiently in craniocaudal and mediolateral oblique view images. The authors compared the binary classification of the two models. The initial model attained an accuracy of 88.1\%, while the subsequent one reached 87.9\%. 

In~\cite{lou2021mgbn} authors proposed a novel multi-level global-guided branch-attention network (MGBN) for mass classification, aiming to optimize feature representation by fully utilizing multi-level global contextual information. The MGBN comprises a stem module and a branch module. The stem module employs standard local convolutional operations from ResNet-50 to extract local information. The branch module, on the other hand, integrates global contextual information and establishes relationships across different feature levels using global pooling and Multi-layer Perceptron (MLP). The final prediction is generated by combining both local and global information. To enhance the reliability and interpretability of the mass classification network, coarse localization maps are visualized using Grad-CAM. Visualization results demonstrate that, compared to ResNet and SENet, MGBN can more accurately locate the mass region, a crucial aspect in clinical diagnosis. 
In recent work, the authors~\cite{raghavan2023attention} suggested that classification and segmentation are co-related tasks, and approaches like~\cite{wang2023information} and ~\cite{ren2023uncertainty} proposed multitask approaches to perform these tasks for breast cancer images. While most existing algorithms are single-task models, Wang et al.\cite{wang2023information} propose an interpretable multitask information bottleneck network (MIB-Net) for breast cancer diagnosis. MIB-NET identifies discriminant features and offers a visual interpretation. It operates based on the principle of an information bottleneck, which involves maximizing the information shared between the latent representations and the target labels while minimizing the information shared between the latent representations and the inputs\cite{tishby2015deep}, \cite{schulz2020restricting}. 
The Grad-CAM was employed to accentuate the regions influencing the prediction outcomes, thereby examining the impact of attention mechanisms and information bottleneck on gradients and class activations.

Most studies reported up until now were performed using 2D mammography. \textbf{DBT} has quickly become widely adopted for standard breast cancer screening, demonstrating enhanced screening effectiveness compared to digital mammography (DM)~\cite{lowry2020screening}.  For example, Ricciardi et al.~\cite{ricciardi2021deep} developed a deep CNN-based computerized detection system to classify mass lesions using DBT. 
The study utilizes a binary classification framework employing the architectures of AlexNet and VGG-19 to discern the existence of mass lesions in DBT images from two proprietary datasets. Employing the Grad-CAM method, the authors investigate the classifiers' behavior to determine their alignment with lesion delineations provided by expert radiologists. Through this analysis, the authors infer that the central regions of the lesion play a more significant role in classification, while the tumor's branches exert less influence on classification tasks.

Likewise, Hussain et al.~\cite{hussain2022shape} introduce a deep-learning-driven multiclass shape-based classification framework for tomosynthesis of breast lesion DBT images, offering both mathematical and visual explanations. The authors explored eight pre-trained CNN architectures to classify lesion-containing ROI images. Additionally, the study delves into the opaque nature of deep learning models using two prominent explainable AI algorithms, Grad-CAM and LIME (as detailed in Section \ref{lime}). Additionally, two interpretability techniques based on mathematical structure, namely t-SNE~\cite{van2008visualizing}, and UMAP~\cite{mcinnes2018umap}, are utilized to examine the pre-trained model's response to multiclass feature clustering. The experimental results of the classification task confirm the effectiveness of the proposed framework, achieving a mean area under the curve of 98.2\%.

\subsection{\underline{L}ocal \underline{I}nterpretable \underline{M}odel-agnostic \underline{E}xplanations (LIME)}\label{lime}

Ribeiro et al. ~\cite{ribeiro2016} introduced LIME to explain predictions of complex models like deep neural networks. LIME is a post-hoc, model-agnostic technique that approximates the decision boundary of a complex model as linear in the local vicinity of the instance being explained. It generates perturbed samples around the instance, obtains predictions from the complex model, assigns weights based on proximity to the instance, and builds an interpretable model on these weighted samples. The goal is to provide a locally accurate and understandable explanation for the instance in question, enhancing model transparency.

\textbf{LIME Equation:}
The explanation is derived as:
\begin{equation}\label{eq:5}
\xi(x) = \arg\min_{g \in G} L(f, g, \pi_x) + \Omega(g)
\end{equation}
where, $G$ different explanation families, fidelity functions $L$, and complexity measures $\Omega$, focusing on sparse linear models.

\begin{enumerate}
    \item \textbf{Interpretable Data Representations:}
    Explanations require a human-comprehensible representation, independent of the model's specific features. For text classification, an interpretable representation might be a binary vector indicating word presence or absence. For image classification, it might indicate the presence or absence of a super-pixel. The variable \( x \in \{R\}^d \) represents the original instance, while \( x_0 \in \{0, 1\}^{d_0} \) is its interpretable form.

   \item \textbf{Fidelity-Interpretability Trade-off:}
The explanation model \( g \in G \) should be simple to interpret. The complexity of \( g \) is denoted by \( \Omega(g) \). The model \( f: R^d \rightarrow R \) represents the original classifier, with \( f(x) \) indicating class probability. Proximity \( \pi_x(z) \) defines the locality around \( x \). Discrepancy between \( g \) and \( f \) within this locality is quantified by \( L(f, g, \pi_x) \). The goal is to minimize \( L(f, g, \pi_x) \) while keeping \( \Omega(g) \) low for interoperability.

   \item \textbf{Sampling for Local Exploration:}
To minimize locality-aware loss $L(f, g, \pi_x)$ without assumptions about $f$, a sample is drawn weighted by $\pi_x$. Perturbed samples are generated by randomly selecting non-zero elements of $x_0$, converted back to the original representation, and labeled by $f(z)$. The dataset $Z$ comprising perturbed samples and labels is used to optimize Eq. (\ref{eq:5}). LIME samples instances near and far from $x$, providing resilience to sampling noise due to $\pi_x$ weighting.

     \item \textbf{Sparse Linear Explanations:}
Consider $G$ as linear models, $g(z_0) = w_g \cdot z_0$, and locally weighted square loss $L$ as:
\begin{equation}\label{eq:6}
L(f, g, \pi_x) = \sum_{z, z_0 \in Z} \pi_x(z) (f(z) - g(z_0))^2
\end{equation}
For text classification, the interpretable representation is a bag of words, with a maximum word constraint $K$. For image classification, "super-pixels" are used instead of words. Direct solution of Eq. (\ref{eq:5}) is impractical, so K-LASSO is used.

\begin{equation}\label{eq:7}
\Omega(g) = \infty1[||w_g||_0 > K]
\end{equation}
\end{enumerate}
    
\subsubsection{LIME: A Post-Hoc Explanation Method in Breast Cancer Diagnosis}
Analogous to SHAP and Grad-CAM, researchers have adapted LIME for an explanation of breast cancer classification and detection models, for instance, Lee et al. ~\cite{lee2021axillary} suggest the classification of \textbf{ultrasound images} of lymph nodes to predict nodal metastasis and early breast cancer (achieving an accuracy of 81.05\%). Additionally, authors used LIME to graphically identify regions of interest (ROIs) in individual images. LIME employed the Simple Linear Iterative Clustering (SLIC) superpixel segmentation method to partition the image into small components, referred to as perturbed regions. It then utilized a linear model to determine the relevance of each superpixel image for a given classification. While the parameters of the superpixel method could impact the explained results, in most cases, there is general agreement on which region of the image is focused ~\cite{palatnik2019local}. The authors specifically considered the image region that influences the prediction results of the proposed method; hence, LIME was employed to extract the region with the highest prediction probability from the image.
Deshmukh et al.~\cite{deshmukh2023explainable} introduced an enhanced quantum k-means (\textit{qk}-means) clustering algorithm and assessed its performance using the Breast Cancer dataset. The experiment involved 600 records of BC patients, each characterized by seven features. The authors then conducted calculations, comparisons, and evaluations based on accuracy, completeness, and silhouette score to demonstrate the clustering accuracy and reliability of the improved \textit{qk}-means algorithm. Notably, the \textit{qk}-means clustering algorithm surpassed all existing algorithms, achieving a higher accuracy of 93.2\% and completeness of 94.5\%. The LIME approach and the improved \textit{qk}-means clustering algorithm are merged to explain the predictions. LIME analyzed the observations and calculated the distance metrics or similarities, between the test data sets. The improved \textit{qk}-means algorithm makes clusters for predictors. Finally, LIME tries a combination of predictors, i.e. \textit{n} number of predictors, to figure out the minimum number of predictors that give the maximum likelihood of the clusters predicted by the improved \textit{qk}-means algorithm.

Some researchers have focused on metastatic cancer types along with primary breast cancers as they are responsible for 90\% of cancer-related deaths. In addition to ultrasound images researchers have employed LIME over \textbf{genomic datasets} for instance, Modhukur et al.~\cite{modhukur2021machine} have introduced various classifiers, including XGBoost, SVM, NB, and RF machine learning models for the classification of cancer types based on their tissue of origin. The authors utilized 24 cancer types and 9303 methylome samples obtained from TCGA and the GEO. The experiments indicated that RF surpassed the other classifiers, achieving an average accuracy of 99\%. Moreover, LIME was used to explain important methylation biomarkers to classify cancer types. The authors further recognize the robustness of LIME explanations, indicating that slight alterations in the input feature vector can significantly impact the LIME explanations regarding feature importance indicators and the corresponding weights assigned to each feature in ~\cite{alvarez2018robustness}.
One of the ways to detect breast cancer is to use histopathology by examining cancer cell tissues. However, manual examination conducted by histopathologists for cell profiling is labor-intensive, time-consuming, and demands considerable expertise, making it costly and necessitating years of training. Kaplun et al.~\cite{kaplun2021cancer} presented an automatic breast cancer cell image analysis system on the public BreakHis dataset ~\cite{spanhol2015dataset}. The authors utilized Zernike image moments to extract complex features from cancer cell images and employed simple neural networks for binary classification (benign vs. malignant classes). Later, the LIME explainability technique was employed over the test results. The use of LIME elucidates the test outcomes of input images by spotlighting the crucial regions accountable for the decisions made by the machine learning algorithm.

Furthermore, as discussed in Section \ref{shap}, Zhang et al. ~\cite{zhang2022applications} applied three posthoc XAI models, including SHAP, LIME, and PDP. The results of the LIME model reveal that particularly the features “worst area”, “worst radius”, “worst perimeter”, “worst texture”, “worst concavity”, “mean texture,” and “mean area” have a positive reaction on the prediction. These results coincide with the results of the SHAP model.

\subsection{Other XAI Techniques}\label{other}
Apart from the techniques mentioned earlier, there are other methods within XAI. These encompass methods such as counterfactual explanations, Permutation Importance, and numerous others. These other methods also show potential for making complex machine learning models easier to understand. 

\subsubsection{Visual Case-Based Reasoning (CBR)}\label{cbr}
Case-based reasoning (CBR)~\cite{althoff2001case} is an AI technique adapted from the cognitive science domain. It is a form of analogical reasoning based on a memory-centered cognitive model. CBR is a problem-solving approach capable of utilizing the specific knowledge of historical cases~\cite{aamodt1994case} to solve similar new cases (aka query cases). In CBR terminology, a case refers to a problem situation. In general, CBR comprises of a \((p+1)\)-dimensional space case database represented as \(X\), where \(x_i\) represents each case. Each \(x_i\) has components in different dimension spaces \(\mathcal{A}_1, \ldots, \mathcal{A}_k, \ldots, \mathcal{A}_p\) and \(Y\) is the solution space associated with each case. The query case can be represented as \(q \in 
 \mathcal{A}_1 \times \ldots \times \mathcal{A}_k \times \ldots \times \mathcal{A}_p\)
Typically, CBR consists of four key steps; retrieve, reuse, revise, and retain. 
The first retrieval step identifies the relevant cases from the database of cases similar to the existing problem. The second step reuses the solutions from the retrieved cases to the current problem. This can also involve adapting the solutions to fit the new context. The third step performs revisions if required, to the proposed solution based on the particulars of the current problem. The final step stores the solution to the new problem and the additional information gained from the experience in the case base for future use. CBR is widely used in different domains, including medical diagnosis, customer support, troubleshooting, legal reasoning, etc.

CBR is also used in the literature to fulfill the need for explainability in breast cancer detection. One of the initial studies was proposed by Lamy et al.~\cite{lamy2019hierarchical} and later an extended version was presented in~\cite{lamy2019explainable} in 2019. 
The authors of this article~\cite{lamy2019explainable} introduced a visual CBR approach for XAI in breast cancer. CBR conducts medical assessments, such as classifications, by comparing a query case (new data) with analogous existing cases from a database. Lamy et al.~\cite{lamy2019explainable} integrate CBR with an algorithm that visually illustrates the similarity between cases, offering users proxies and metrics to interpret. By examining these proxies, users can determine whether to adopt the algorithm's recommended decision.
The article also asserts that medical experts appreciate the visual information with a clear decision-support system.
The proposed hierarchical visual CBR approach can support complex decision-making, such as breast cancer therapy, by dividing the decision process into several, simpler decisions. This work's perspective is verifying the system, possibly using a systematic approach in combination with ML for extracting rules and comparing them with guidelines.
Nevertheless, CBR has some limitations, including its low accuracy and sensitivity to the high dimensionality of attributes.

In another study, Gu et al.~\cite{gu2020case} proposed a method that integrates CBR and ensemble learning for predicting breast cancer recurrence. The model employed a case-based explanation approach to elucidate the output predictions of XGBoost, thereby aiding clinicians in making informed decisions. The predictive ensemble model is constructed using the XGBoost algorithm, while the interpretation of the predictions is facilitated by the CBR approach. This method represents an advancement in current machine learning research, which has primarily prioritized enhancing accuracy over-delivering pertinent explanations for the predictions.
This study has some limitations. First, it is designed within the framework of a system and adopts existing methods. Second, there is a need to investigate
more important attributes that influence the risk of breast cancer recurrence,
taking into account the evolution of cancer and the care relationship between caregivers and patients. Third, to improve the readability of the system's interpretation results, make them more consistent with the habits of doctors, and provide convenience for doctors.

\subsubsection{Permutation Importance (PIMP)}\label{pimp}
PIMP initially proposed in ~\cite{altmann2010permutation} identifies the most significant features. It computes the importance of the features by shuffling their values. A feature is deemed significant if shuffling it leads to a notable error in the model's prediction.
As discussed in Section \ref{shap}, Khater et al.~\cite{khater2023explainable} employed three model-agnostic explainable artificial intelligence (XAI) approaches, namely SHAP, PDP, and PIMP. PIMP highlighted the significance of two key features: Bare nuclei (similar to SHAP) and clump thickness, based on the feature importance scores. These scores are determined by measuring the reduction in a model's performance when a specific feature is randomly permuted. Additionally, PDP interpretations were utilized to corroborate the results of the permutation analysis, as discussed in the subsequent subsection.

\subsubsection{Partial Dependence Plot (PDP)}\label{pdp}
PDP is a global interpretation method. It helps answer "how" the identified features could change the prediction, unlike PIMP which could discover "what" are the most important features only ~\cite{gianfagna2021explainable}. 
As discussed in Subsection \ref{pimp}, PIMP identified the most important features ~\cite{khater2023explainable} i.e. Bare nuclei and clump thickness. Furthermore, the authors developed PDP to discover the relationship feature holds with the predicted output. PDP assists in comprehending the potential correlation between alterations in tissue architecture and the progression of breast cancer by unveiling the relationship between tissue thickness and cell stickiness.

In a different study, Zhang et al.~\cite{zhang2022applications} employed PDP to explain a black-box model by visualizing the influence of subsets of features on the model's predictions, alongside SHAP and LIME.

\subsubsection{Triaged Region of Interest }\label{roi}
Tosun et al.~\cite{tosun2020explainable} outlined the development of an initial Explainable AI (XAI)-enabled software application called HistoMapr-Breast, designed for breast core biopsies. HistoMapr-Breast automatically visualizes breast core whole slide images (WSI) and identifies regions of interest (ROI), presenting key diagnostic areas interactively and explainably. The software offers a reliable explanation interface for pathologists, integrating the concept of XAI into pathology workflows to enhance safety, reliability, and accountability. This integration addresses issues related to bias, transparency, safety, and causality. It's important to note that the XAI system complements pathologists' work rather than replacing it.

\subsubsection{Layerwise Relevance Propagation (LRP)}\label{lrp}
LRP~\cite{bach2015pixel} is a technique used for the “decomposition of nonlinear classifiers” to interpret complex DNNs by spreading their forecasts backward. LRP starts from the output neurons backpropagating to the input-layer neurons. Every neuron distributes an equal amount of information to the lower layer as it acquires from the higher layer. 
Binder et al.~\cite{binder2021morphological} employed the LRP for identifying image regions/pixels that contribute the most to machine-learning-based predictions. LRP is used to highlight molecular biomarkers such as gene expressions that have been inferred from whole slide images (WSIs), and eventually linked with the prognosis of breast cancer. 
A bag-of-words model is suggested for the prediction task, incorporating invariances to rotation, shift, and scale variations in the input data. To validate the prediction outcomes, the Layer-wise Relevance Propagation (LRP) technique is employed in this model. This technique generates heatmaps, providing per-pixel scores that highlight the presence of tumorous structures.
Moreover, LRP heatmaps computed for various target cell types can be amalgamated to generate computationally predicted fluorescence images. These explanations offer histopathologically meaningful insights and may potentially furnish valuable information about which tissue components are predominantly indicative of cancer.
Similarly, Cherada et al.\cite{chereda2021explaining} also advocated for the use of Graph LRP (GLRP) to explain the decisions made by deep learning models. In their study, the authors employed Graph-Convolutional Neural Networks (Graph-CNN) on structured gene expression data to forecast metastatic events in breast cancer. The research utilized gene expression profiles organized by protein-protein interactions (PPI) sourced from the Human Protein Reference Database (HPRD)\cite{keshava2009human}. Subsequently, LRP~\cite{bach2015pixel} was applied to pinpoint and scrutinize the biological relevance of the most influential genes for predictions~\cite{chereda2021explaining}. Pathway analysis of these genes revealed their inclusion of oncogenes, molecular subtype-specific genes, and therapeutically targetable genes such as EGFR and ESR1.
\subsubsection{Adaptive Dimension Reduction}\label{adr}

Amoroso et. al.~\cite{amoroso2021roadmap}, suggested an XAI method based on adaptive dimension reduction for breast cancer therapies. The authors applied the clustering and dimension reduction method, and the experiment results demonstrated that the framework could outline the most important clinical features for the patient and design oncological therapies. 

\subsubsection{Counterfactual}\label{counterfactual}

In recent times, Counterfactual explanations have been recognized as a significant post-hoc method that offers compelling insights for users to comprehend the internal workings of AI models~\cite{biran2017explanation, byrne2019counterfactuals, guidotti2019factual, miller2019explanation, wachter2017counterfactual}. Hanis et al.~\cite{hanis2022over} investigated the influence of various machine learning models on the counterfactual generation process. The authors focused on diagnosing the likelihood of occurrence and predicting the accuracy of breast cancer patients based on their medical history details retrieved from the medical registry to facilitate early treatment. Essential risk factors such as family history, marital status, age group, medical history, and demographic location were considered for occurrence conditions. Additionally, factors related to the breast area, including nipple retraction, discharge, and skin color changes, were taken into account for tumor development. Eight different ML classifier methods were employed to classify and extract features for predicting accuracy factors, scores, etc. The results demonstrated that KNN exhibited a good accuracy F-score compared to other models. Later, the authors utilized a model-agnostic approach to improve the explainability of the proposed system.

\subsubsection{Feature Importance Measures}\label{feature importance}
The presence of a high number of input features prevents the ML model's performance. Hence, some of the prominent explainability techniques focus on the concept of feature importance. It explains a previously constructed model as a set of important features and their respective importance. Generally, these explanations represent features that are interpretable to the users. In ~\cite{saarela2021comparison} have compared different classification explanation models, including feature importance measures ~\cite{bhatt2020explainable}. This research article addresses the following research questions: What are the most important features? 
Do the primary features vary across different techniques, and if they do, which techniques should be deemed reliable? Furthermore, under what circumstances can local explanations enhance the overall understanding provided by global modular explanations of a model, and should they be included in medical studies? The feature importance measures are compared over publicly available breast cancer data from UCI.

\subsection{No XAI Method Defined}\label{mixed}
There are certain approaches in which no specific XAI algorithm is mentioned.
Idrees et al.~\cite{idrees2022explainable}, proposed the XAI algorithms to analyze the potential \textbf{biomarkers} in breast cancer. Different classification models were applied to anthropometric data from three studies~\cite{santillan2013tetrad} ~\cite{assiri2016evaluation} ~\cite{patricio2018using} augmented with XAI models. The outcomes of the XAI technique reveal that the levels of \texttt{HOMA}, \texttt{leptin}, \texttt{adiponectin}, and \texttt{resistin} in serum can serve as novel biomarkers for breast cancer, demonstrating sensitivity ranging from 83 to 87\%, specificity ranging from 81 to 88\%, and AUC ranging from 0.82 to 0.89 with a 95\% confidence interval. These findings are promising for integrating the characteristics of \texttt{HOMA}, \texttt{leptin}, \texttt{adiponectin}, and \texttt{resistin}, potentially facilitating the development of a cost-effective breast cancer biomarker.

This article is ~\cite{kabakcci2021automated} not directly related, so we can add a little detail if needed. This article is about immunohistochemistry and xai for breast cancer tissue images.
This article presents a new automated method for scoring CerbB2/HER2 receptors in breast cancer tissue images using histogram-based analysis. The method aims to replace the manual scoring done by pathologists, which is laborious and prone to error. The proposed approach involves separating the tissue images into color channels, segmenting cell nuclei and boundaries, and extracting features representing the intensity and completeness of membrane staining. These features are subsequently utilized to calculate cell-based HER2 scores, which are then combined following ASCO/CAP recommendations to derive the final tissue score. The efficacy of the method is assessed using two publicly available datasets and compared against state-of-the-art techniques. The findings demonstrate that the proposed method is highly effective in scoring HER2 tissue on both balanced and unbalanced datasets, offering a transparent and interpretable approach for HER2 tissue scoring. Various classifiers are tested in this study. For the ITU-MED datasets, the highest ranking, both for cell-based and tissue-based scoring, is an Ensemble Boosted Tree, achieving 77.56\%/91.43\% (cell/tissue) accuracy on ITU-MED-1 and 91.62\%/90.19\% on ITU-MED-2. For the contest dataset, a Cosine kNN achieved the highest results, ranking 6th among all 18 contest competitors.

 \section{Discussion} \label{discussion}

\begin{figure}[hbt!]
\includegraphics[width=16cm, height=4cm]{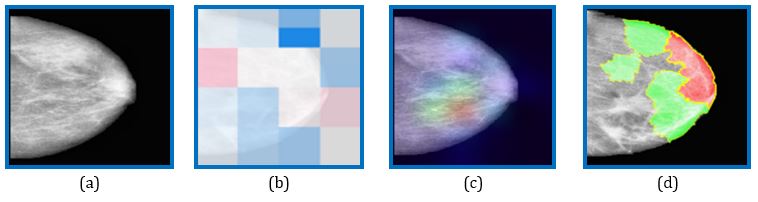}
\caption{Comparative visualization of XAI techniques for breast cancer detection~\cite{ahmed2024enhancing}. (a) The input image (b) SHAP (c) Grad-CAM (d) LIME}
\label{fig16}
\end{figure}

In this section, we discuss the three most popularly used XAI models in the breast cancer domain, including SHAP, Grad-CAM, and LIME, and their visualization is presented in Figure~\ref{fig16}, labeled as \textbf{b,c, and d} respectively.

To begin with, SHAP is "model-agnostic," meaning it is not tied to any specific ML/DL model. This is beneficial because it allows SHAP values to be used with any machine learning model, regardless of its architecture or underlying algorithms~\cite{adadi2018peeking}. This flexibility is useful for tasks where different models can be compared, as it enables a consistent method for interpreting their predictions and understanding their behavior. Since SHAP values are independent of the specific model being used, they can be easily adapted when the model itself is updated or changed. In a machine learning pipeline, where models may need to be refined or replaced over time, this flexibility ensures that the explanations provided by SHAP values remain consistent and informative, requiring minimal effort to integrate with updated models~\cite{ribeiro2016model}.

Shapley values also offer three useful properties that are not found in other XAI methods~\cite{lundberg2017},~\cite{lundberg2018explainable}. First, SHAP values provide local accuracy, meaning they accurately capture the difference between the expected output of the model and the output for a specific instance of input data. This means they provide insights into how individual features contribute to each prediction. Second, if a feature is missing (i.e., not present in the input data), its attribution value in the SHAP analysis is set to zero, ensuring that the absence of a feature is appropriately accounted for in the explanation. Third, changes in the model should not result in contradictory changes to the SHAP values assigned to individual features. If increasing the importance of a feature in the model's prediction also increases its corresponding SHAP value, then consistency is maintained. These advantageous properties make the SHAP model an ideal choice for different research studies' objectives. It offers a robust and reliable method for explaining model predictions and comparing different models, regardless of their type or complexity. For example, SHAP performed well with the XGBoost model, as witnessed in ~\cite{moncada2021explainable},~\cite{massafra2023analyzing},~\cite{vrdoljak2023applying},~\cite{silva2023hybrid}, and~\cite{khater2023explainable}.

Despite its benefits, the SHAP method may have limitations. For instance, the computational burden of computing SHAP values for certain "generic" models can be significant, potentially slowing down the analysis process. Additionally, while SHAP values are optimized for some model architectures, they may not perform as efficiently as others. In scenarios of constrained computational resources or when real-time analysis is necessary, these constraints could impact the practicality of using SHAP values. While SHAP values are valuable, they are not a universal solution for all explainability needs. Depending on the specific application or context, other emerging explainability techniques may be more suitable or offer complementary insights. Thus, it's essential to consider the specific requirements and constraints of each scenario when choosing an explainability method~\cite{guidotti2018survey}~\cite{lamy2019explainable}.

Grad-CAM is a widely applied XAI method in medical images~\cite{chaddad2023survey}. It generates heatmaps that highlight regions of an image that contribute significantly to the model's decision, providing insights into which features or areas the model focuses on for classification. This interpretability aids radiologists or clinicians in understanding and validating the model's predictions. The research studies discussed in this survey applied Grad-CAM to predict whether an image contains cancerous features and localize the specific areas in the image that contribute to the prediction. This localization capability can help pinpoint the location of abnormalities within breast tissue, potentially improving diagnostic accuracy. Techniques such as~\cite{masud2020convolutional},~\cite{suh2020automated},~\cite{dong2021one}, and~\cite{lou2021mgbn} indicated that Grad-CAM serves as a valuable tool to corroborate the findings of deep learning models by providing visual evidence of the features that drive the predictions.

However, one of the shortcomings of Grad-CAM is that it relies on the gradient information computed during backpropagation, which limits its applicability to CNN-based models. If the breast cancer diagnosis model is not based on CNNs, then it cannot be directly applied. Additionally, Grad-CAM highlights regions of high importance for classification, but it may prioritize features that are not clinically relevant or meaningful for diagnosis. This could lead to false positives or misinterpretations of the model's decisions if salient features do not correspond to actual cancerous lesions. Additionally, the Grad-CAM techniques generate a rough localization map. It is noted that, at times, XAI methods explain the outcomes of DL methods using slightly divergent regions. This discrepancy may arise due to model overfitting~\cite{hussain2022shape}. A thorough investigation could yield insightful conclusions. The CAM technique concentrates solely on a broad area within an image, neglecting subtle nuances, unlike the LIME technique, which creates perturbations and emphasizes key features. Likewise, the SHAP model precisely measures the contribution of specific regions and holds promise for inclusion in future research endeavors.

Raghavan et al.~\cite{raghavan2023attention} proposed Attention-guided Grad-CAM (AGG-CAM), which generates improved visual representations by accounting not only for the gradient importance of individual pixels but also for the significance of various input components determined by the attention mechanism. This yields heatmaps that are more interpretable, resilient, and targeted compared to conventional Grad-CAM approaches. These findings offer a mathematical insight into how attention mechanisms can refine and augment Grad-CAM visualizations. The attention-guided Grad-CAM method outlined in this article presents a hopeful avenue for improving the interpretability and transparency of deep learning models. Implementing this technique in clinical settings has the potential to transform breast cancer detection and positively improve patient outcomes. In conclusion, while Grad-CAM offers valuable insights into the decision-making process of deep learning models for breast cancer diagnosis, its applicability, reliability, and interpretability should be carefully considered in conjunction with other diagnostic methods and clinical expertise.

In resemblance with SHAP, LIME is also model-agnostic, meaning it can be applied to any machine learning model, including those used for breast cancer diagnosis. This flexibility allows it to be used with various models without modifications. It provides explanations at the individual prediction level, highlighting the features that contributed most to a specific prediction. This local interpretability is crucial in medical diagnosis, as it helps clinicians understand why a particular prediction was made and assess its reliability. Though LIME explanations are locally faithful, explanations might not be reasonable at the global level, as discussed in~\cite{deshmukh2023explainable}. Additionally, LIME explanations lack robustness. Minor alterations in the input feature vector can significantly impact LIME explanations, affecting the indications of feature importance and the assigned weights for each feature~\cite{alvarez2018robustness}. LIME can validate machine learning models' decisions by providing interpretable explanations for their predictions. Clinicians can compare the model's explanations with their domain knowledge and diagnostic findings to assess the model's reliability and identify potential errors or biases.

The integration of these XAI techniques into clinical workflows holds great promise for improving the transparency and reliability of ML/DL models in breast cancer diagnosis. Future research should focus on developing more efficient algorithms for XAI, standardizing evaluation metrics to assess the quality of explanations, and exploring the integration of multiple XAI techniques to leverage their complementary strengths. Furthermore, the impact of XAI on user trust and comprehension should be systematically studied to ensure that these techniques not only enhance model interpretability but also foster confidence among medical practitioners and patients. As XAI continues to evolve, its potential to revolutionize personalized medicine and improve patient outcomes in breast cancer diagnosis becomes increasingly evident.

\section{Conclusion}\label{conc}

Despite the widespread use of machine learning and deep learning black-box models for breast cancer classification, there is a notable lack of in-depth studies focused on individual XAI techniques. This gap underscores the necessity for future research to explore these techniques more thoroughly, providing unique insights into the complex mechanisms of ML/DL models. Enhancing the transparency and trustworthiness of these models is crucial.

XAI plays a vital role in evaluating model performance, which is essential for optimizing models for greater accuracy and reliability in practical applications. Beyond performance, it is imperative to assess the impact of XAI on user trust and comprehension. As AI technologies become increasingly integrated into the healthcare sector, ensuring that these technologies are both transparent and interpretable is essential for fostering trust among medical practitioners and patients alike. Integrating various XAI approaches presents significant challenges, particularly in their comparative analysis and the generalization of their efficacy across different scenarios. This complexity highlights the need for a robust framework to evaluate these models comprehensively. Establishing standardized evaluation metrics is critical, as they provide consistent and reliable measures of effectiveness in clinical settings. These metrics should not only assess the accuracy and efficiency of the XAI applications but also evaluate the quality of the explanations in terms of their transparency, relevance, and user understandability.

When evaluating XAI models, it is crucial to establish appropriate evaluation metrics to quantify and qualify the explanations provided by the models. These metrics should reflect the quality of the explanation, including factors such as relevance, completeness, and consistency. By developing standardized metrics for evaluating XAI models, researchers and practitioners can ensure that the models are transparent and interpretable, which is critical for gaining the trust of both practitioners and patients. Additionally, the integration of XAI can lead to advancements in personalized medicine. By offering tailored explanations, XAI can help clinicians understand why certain diagnostic or treatment decisions are recommended. This is particularly important in complex cases such as breast cancer, where patient-specific factors can significantly influence outcomes. The potential for XAI to allow for more nuanced and customized healthcare interventions could dramatically improve patient care and treatment efficacy.

However, AI has the potential to revolutionize breast cancer detection and treatment, but significant challenges remain. These include exploring XAI techniques, developing standardized metrics for evaluation, and integrating these technologies more deeply into clinical practice. As research in this exciting field continues to progress, it is essential to focus on developing models that are not only effective but also user-friendly, understandable, and trustworthy. The successful implementation of XAI could have a profound impact on the future of healthcare, making it a critical area for ongoing research and development.

\newpage
\bibliographystyle{IEEEtran}




\end{document}